\documentclass[11pt]{article}
\usepackage[margin=1in]{geometry}
\usepackage{graphicx}
\usepackage{amsmath,amssymb,amsfonts}
\usepackage{booktabs}
\usepackage{multirow}
\usepackage[super,sort&compress]{natbib}

\usepackage{lineno}
\usepackage{setspace}
\usepackage[hidelinks]{hyperref}
\usepackage{xcolor}
\usepackage{textcomp}
\usepackage[title]{appendix}
\usepackage{algorithm}
\usepackage{algorithmicx}
\usepackage{algpseudocode}
\usepackage{listings}
\usepackage{authblk}
\usepackage{xcolor}
\usepackage{hyperref}
\hypersetup{
    colorlinks=true,
    linkcolor=blue,
    citecolor=black,
    urlcolor=blue
}
 
\usepackage{graphicx}%
\usepackage{multirow}%
\usepackage{amsmath,amssymb,amsfonts}%
\usepackage{amsthm}%
\usepackage{mathrsfs}%
\usepackage[title]{appendix}%
\usepackage{xcolor}%
\usepackage{textcomp}%
\usepackage{manyfoot}%
\usepackage{booktabs}%
\usepackage{algorithm}%
\usepackage{algorithmicx}%
\usepackage{algpseudocode}%
\usepackage{listings}%

\title{A Unified Vision-Language Model for PSMA PET/CT Report Generation, Visual Question Answering, and Lesion Segmentation}

\author[1]{Yang Xing}
\author[1]{Jiong Wu}
\author[2]{Savas Ozdemir}
\author[1]{Yang Zhou}
\author[1]{Boxiao Yu}
\author[3]{Ying Zhang}
\author[4]{Zheren Zhu}
\author[5]{Chenyu You}
\author[6]{Wei Shao}
\author[7]{Yang Lu}
\author[4]{Kang Wang}
\author[7]{Tinsu Pan}
\author[4]{Yang Yang}
\author[1]{Kuang Gong\thanks{Correspondence: kgong@bme.ufl.edu}}

\affil[1]{J. Crayton Pruitt Family Department of Biomedical Engineering,
University of Florida, Gainesville, FL, USA}

\affil[2]{Department of Radiology,
University of Florida, Jacksonville, FL, USA}

\affil[3]{Department of Research Computing,
University of Florida, Gainesville, FL, USA}

\affil[4]{Department of Radiology,
University of California, San Francisco, San Francisco, CA, USA}

\affil[5]{Department of Applied Mathematics \& Statistics,
Stony Brook University, Stony Brook, NY, USA}

\affil[6]{Department of Medicine,
University of Florida, Gainesville, FL, USA}

\affil[7]{Division of Diagnostic Imaging,
The University of Texas MD Anderson Cancer Center, Houston, TX, USA}

\date{}
\begin{document}
\maketitle


% \abstract{Accurate PSMA PET/CT interpretation is central to prostate cancer management, yet existing PET/CT AI models typically address isolated tasks. We propose a unified PSMA PET/CT vision-language model for report generation, visual question answering, and lesion segmentation. The framework adopts an LLaVA-style architecture, comprising a PET/CT vision encoder, an MLP-Mixer projection module, a LoRA-tuned large language model, and a 3D segmentation branch. Training followed a four-stage strategy: vision encoder pretraining, projection-layer alignment, VLM fine-tuning, and final multitask tuning. Language tasks used 5,747 PSMA PET/CT datasets with paired reports, while segmentation used the PSMA subset of AutoPET. The model outperformed PET2REP and a CT-based baseline across standard report-generation metrics, improved performance across VQA question types, and achieved higher Dice and lesion-level overlap F1 than SegAnyPET and nnUNet. These results support the feasibility of a unified framework for structured, interactive, interpretable PSMA PET/CT analysis with voxel-level grounding within a single multitask model architecture.}

\begin{abstract}{Accurate PSMA PET/CT interpretation is central to prostate cancer management, yet existing PET/CT AI models typically address isolated tasks. We propose a unified PSMA PET/CT vision-language model for report generation, visual question answering, and lesion segmentation. The framework adopts an LLaVA-style architecture, comprising a PET/CT vision encoder, an MLP-Mixer projection module, a LoRA-tuned large language model, and a 3D segmentation branch. Training followed a four-stage strategy: vision encoder pretraining, projection-layer alignment, VLM fine-tuning, and final multitask tuning. Language tasks used 5,747 PSMA PET/CT datasets with paired reports, while segmentation used the PSMA subset of AutoPET. The model outperformed PET2REP and a CT-based baseline across standard report-generation metrics, improved performance across VQA question types, and achieved higher Dice and lesion-level overlap F1 than SegAnyPET and nnUNet. These results support the feasibility of a unified framework for structured, interactive, interpretable PSMA PET/CT analysis with voxel-level grounding within a single multitask model architecture.}
\end{abstract}

%%================================%%

% \keywords{Deep learning; PSMA PET; vision-language model; segmentation}


% \maketitle

\section{Introduction}\label{sec1}

Following regulatory approval and rapid clinical adoption, prostate-specific membrane antigen (PSMA) PET has become an essential tool for prostate cancer (PCa) management. It outperforms CT and bone scanning in detecting pelvic lymph nodes and distant metastases~\cite{hofman_prostate-specific_2020, hope_diagnostic_2021}, and may also facilitate intraprostatic tumor localization and contouring~\cite{chen_combination_2019,bettermann_68ga-psma-11_2019}. For PCa biochemical recurrence monitoring, PSMA PET has been demonstrated to be substantially more sensitive than other imaging modalities, especially for prostate-specific antigen (PSA) levels $<$ 1 ng/mL~\cite{cornford_eau-eanm-estro-esur-siog_2021, morigi_prospective_2015, afshar-oromieh_comparison_2014}. PSMA PET is now routinely used for baseline staging, detection of biochemical recurrence, and restaging after therapy. Additionally, quantitative PSMA PET imaging plays a central role in treatment response and therapy eligibility for targeted radioligand therapies such as 177Lu-PSMA~\cite{eiber_evaluation_2015, rauscher_efficacy_2018}. Consensus guidelines~\cite{kratochwil_joint_2023, hope_snmmi_2023, hofman_177lu-psma-617_2018} increasingly emphasize PSMA PET-derived whole-body tumor burden~\cite{sartor_lutetium-177psma-617_2021, gafita_novel_2022},  lesion distribution~\cite{eiber_prostate_2018}, and response patterns~\cite{fanti_proposal_2020} as key determinants of patient outcomes and suitability for targeted radioligand therapies. Therefore, PSMA PET/CT interpretation requires not only visual lesion detection, but also reproducible lesion localization, quantitative uptake assessment, and whole-body disease burden estimation. 

\begin{figure}[htb]
\includegraphics[trim=0cm 2.6cm 0cm -1cm, clip, width=\textwidth]{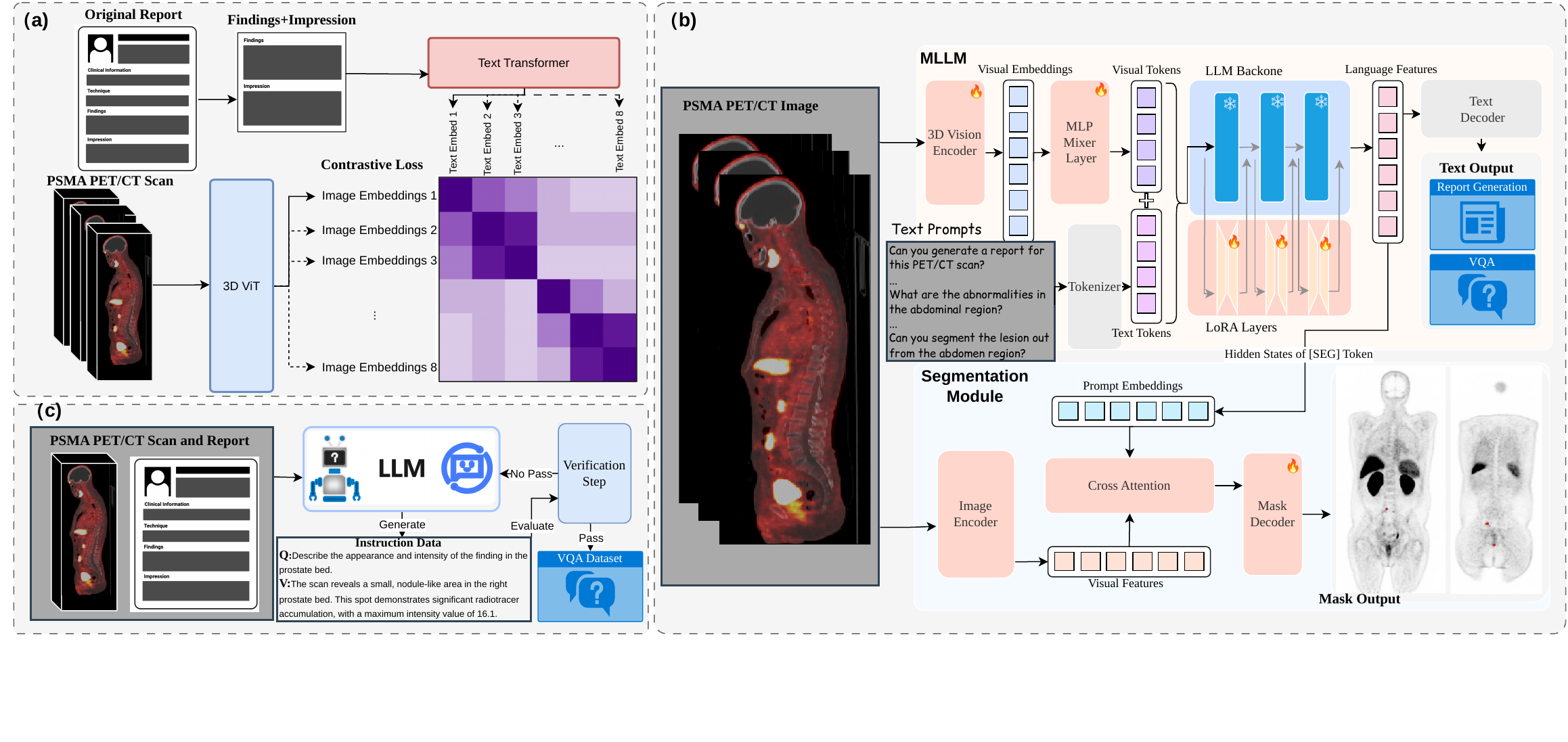}
\caption{Illustration of the proposed model architecture: (a) The architecture of the PET/CT vision encoder. Contrastive loss was used to train the vision and text encoders to pair radiology reports with the corresponding PET/CT scans. (b) An overview of the proposed methods, which consists of a LLaVA-like MLLM and an nnUNet-based segmentation module, connected by passing the hidden states of [SEG] Token from the MLLM to the segmentation model. (c) The curation process of generating VQA pairs using Gemini-2.5-pro. Details of each module are given in Sec.~\ref{sec4}.}
\label{model_architecture}
\end{figure}
Clinical interpretation and reporting of PET imaging remain largely manual and highly dependent on reader expertise. This limitation is particularly evident in PSMA PET imaging, which has been rapidly adopted in clinical practice, yet standardized reporting remains inconsistently applied. Although professional societies have advocated for structured PSMA PET reporting frameworks, a recent study reported that only 11\% of clinical PSMA PET reports adhered to a standardized format, highlighting substantial heterogeneity and inconsistency in real-world practice~\cite{vargas_psma_2025}. Such variability may contribute to misinterpretation, inconsistent follow-up or work-up recommendations, and delays in patient management. These challenges motivate the development of automated frameworks for PSMA PET interpretation and reporting. Vision-language models (VLMs), building on architectures such as CLIP~\cite{radford_learning_nodate} and LLaVA~\cite{liu_visual_2023}, provide a powerful framework for integrating visual and textual information and have been widely applied to medical image analysis, including classification\cite{huemann_contextual_2026}, image-text retrieval\cite{zhang2025biomedclipmultimodalbiomedicalfoundation}, report generation\cite{hamamci_generalist_2026}, visual question answering (VQA)~\cite{wu_radfm_2025,bai2024m3dadvancing3dmedical}, and interactive annotation\cite{zhao2025foundation}. Compared with the growing number of CT- and MR-oriented medical VLMs, PET-specific VLMs remain relatively underexplored. Existing PET VLMs~\cite{zhang_pet2rep_2025, jiao_vision-language_2025, maqbool_petar_2025} have shown potential for image-language alignment and instruction-following, but they have not been designed to jointly support PSMA PET/CT clinical language tasks and 3D lesion segmentation within a single framework.

% Current PET/CT AI methods and medical VLMs have largely been developed for different purposes. Task-specific PET/CT models typically address individual image-analysis objectives, including lesion detection~\cite{huemann_contextual_2026}, segmentation~\cite{zhang_seganypet_nodate}, quantitative biomarker extraction~\cite{fritsak2026glowfdggeneralizedcancerlesion}, and treatment-response assessment~\cite{jemaa2024automated}. Medical VLMs, by contrast, have primarily focused on image--text alignment~\cite{zhang2025biomedclipmultimodalbiomedicalfoundation}, report generation~\cite{hamamci_generalist_2026}, and visual question answering (VQA)~\cite{wu_radfm_2025,bai2024m3dadvancing3dmedical}. Consequently, there remains a need for a unified PSMA PET/CT framework that supports language-based clinical interpretation and explicit 3D lesion segmentation as complementary tasks.

In this study,  we develop a unified PSMA PET/CT VLM for report generation, VQA, semantic lesion segmentation, and language-guided lesion segmentation.  The proposed framework combines clinical language tasks and voxel-level lesion segmentation within a single instruction-driven architecture, allowing the model to generate text responses and produce lesion masks according to task-specific prompts. To enable language-guided segmentation, we adapt a LISA-like~\cite{lai_lisa_2024} language-guided segmentation interface to the VLM framework and replace the generic segmentation decoder with an nnUNet-style 3D lesion segmentation module. Because anatomical context is important for distinguishing physiologic PSMA uptake from suspicious lesions, we incorporate TotalSegmentator-derived organ masks as auxiliary anatomical supervision. The model is trained using image-report pairs for report generation, report-derived question-answer pairs for VQA, and lesion-mask annotations for voxel-level segmentation. Through this design, the proposed framework supports both clinical text generation and explicit lesion segmentation for whole-body PSMA PET/CT imaging. Fig.~\ref{model_architecture} illustrates the architecture of the proposed unified PSMA PET/CT VLM.

\section{Results}\label{sec2}
\textcolor{black}{For image-text retrieval, report generation and VQA tasks, we included 5,747 whole-body PSMA PET/CT image-report paired studies acquired with tracer 18F-DCFPyL between 2022 and 2025. For lesion segmentation task, we utilized PSMA subset from the autoPET challenge~\cite{dexl_autopet3_2026}, containing 597 whole-body PSMA PET/CT image-mask paired studies from 378 male patients acquired at LMU University Hospital Munich between 2014 and 2022.}
\subsection{PET/CT image-text retrieval}
Image-to-text and text-to-image retrieval were used to evaluate the image-text alignment learned by the PET/CT vision encoder, referred to as PET/CT-CLIP. In image-to-text retrieval, a PET/CT scan was used as the query to retrieve its corresponding report or text description, and in text-to-image retrieval, a report or text query was used to retrieve the matched PET/CT scan. Retrieval performance was assessed using Recall@K, defined as the percentage of queries for which the matched item was retrieved within the top K results. The proposed PET/CT-CLIP model was compared with random retrieval and CT-CLIP (14), which served as a CT-based vision-language baseline. As shown in Table~\ref{tab:image_text_retrieval}, PET/CT-CLIP achieved the highest Recall@K values across all evaluated retrieval settings. For image-to-text retrieval, PET/CT-CLIP improved Recall@100 from 14.17\% with CT-CLIP to 35.57\%. For text-to-image retrieval, Recall@100 increased from 8.52\% with CT-CLIP to 38.87\%. Similar improvements were observed at lower K values, indicating improved alignment between whole-body PET/CT scans and corresponding text descriptions after PET/CT-specific pretraining.

\begin{table}[t]
\centering
\small
\caption{Performance comparison for PET/CT image-text retrieval.
Recall@$K$ is reported for image-to-text and text-to-image retrieval
using random selection, CT-CLIP, and the proposed PET/CT-CLIP model.
Values are reported as mean [95\% CI].}
\label{tab:image_text_retrieval}

\renewcommand{\arraystretch}{1.15}
\setlength{\tabcolsep}{4pt}

\begin{tabular}{llccc}
\toprule
\textbf{Task} &
\textbf{Metric} &
\textbf{Random} &
\textbf{CT-CLIP} &
\shortstack{\textbf{PET/CT-CLIP}\\\textbf{(Proposed)}} \\
\midrule

\multirow{4}{*}{Image-to-text}
& Recall@5
& $0.35\,[0.09,\,0.70]$
& $1.04\,[0.52,\,1.74]$
& $\mathbf{3.13\,[2.17,\,4.17]}$ \\

& Recall@10
& $0.78\,[0.35,\,1.31]$
& $1.83\,[1.13,\,2.61]$
& $\mathbf{5.13\,[3.91,\,6.43]}$ \\

& Recall@50
& $4.52\,[3.30,\,5.65]$
& $7.83\,[6.26,\,9.39]$
& $\mathbf{19.91\,[17.65,\,22.35]}$ \\

& Recall@100
& $7.48\,[6.00,\,9.04]$
& $14.17\,[12.17,\,16.17]$
& $\mathbf{35.57\,[32.61,\,38.26]}$ \\

\midrule

\multirow{4}{*}{Text-to-image}
& Recall@5
& $0.87\,[0.43,\,1.48]$
& $0.43\,[0.09,\,0.87]$
& $\mathbf{2.43\,[1.65,\,3.30]}$ \\

& Recall@10
& $0.43\,[0.09,\,0.87]$
& $0.78\,[0.26,\,1.30]$
& $\mathbf{4.87\,[3.74,\,6.09]}$ \\

& Recall@50
& $5.13\,[3.83,\,6.35]$
& $4.35\,[3.22,\,5.65]$
& $\mathbf{23.83\,[21.30,\,26.17]}$ \\

& Recall@100
& $8.26\,[6.78,\,9.74]$
& $8.52\,[6.87,\,10.17]$
& $\mathbf{38.87\,[36.08,\,41.83]}$ \\

\bottomrule
\end{tabular}

\end{table}

\subsection{Report generation}
\begin{figure}[htb]
\includegraphics[trim=0cm 0cm 0cm -1cm, clip, width=\textwidth]{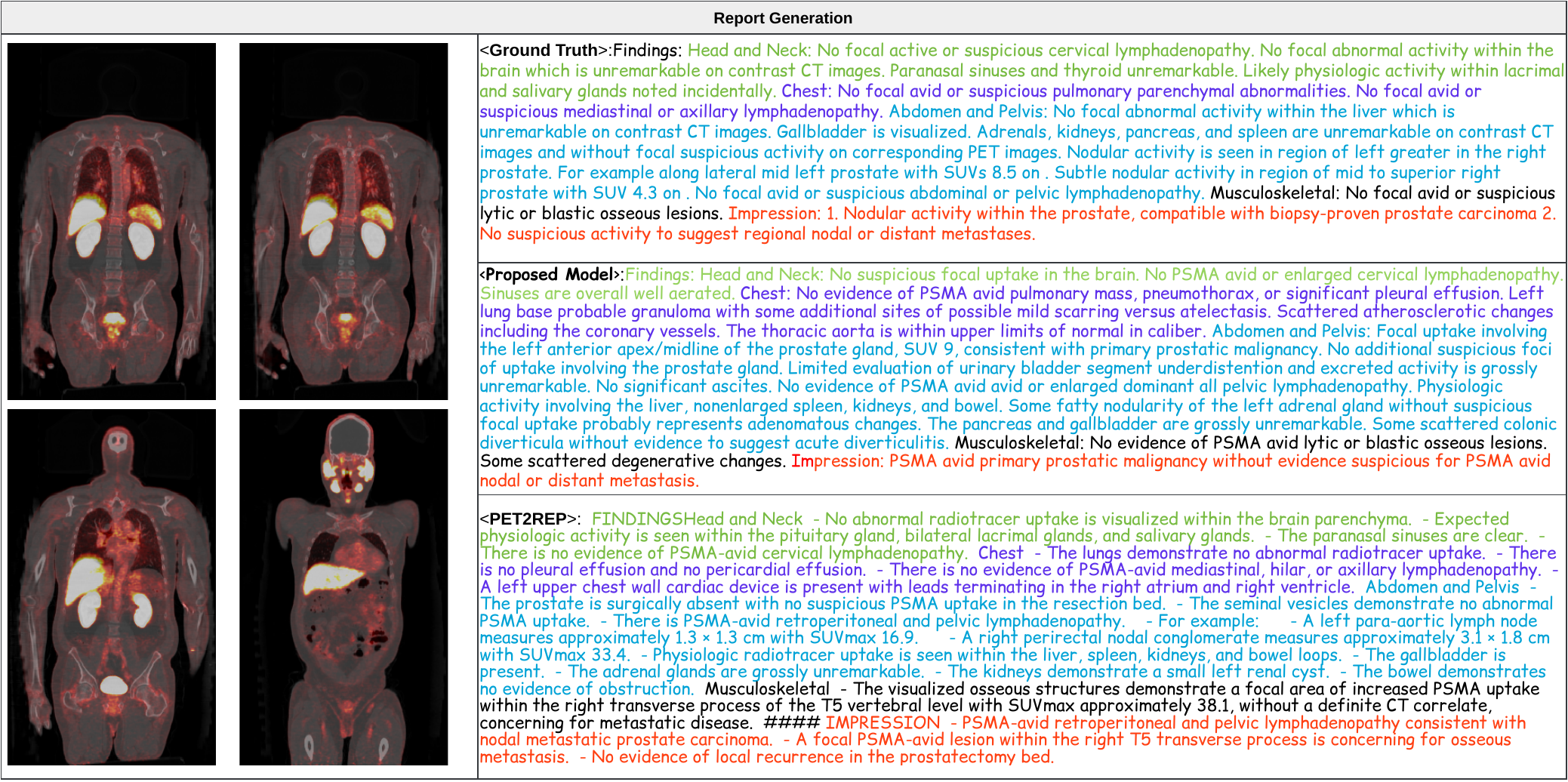}
\caption{Report-generation examples with color-coded report sections. Representative coronal slices from fused PET/CT images are displayed on the left.} 
\label{report_generation_fig}
\end{figure}

Radiology report generation was evaluated on the held-out PSMA PET/CT test set using ROUGE~\cite{lin_rouge_nodate}, BLEU~\cite{papineni_bleu_2001}, METEOR~\cite{satanjeevsatanjeev_banerjee_meteor_2005}, CIDEr~\cite{vedantam_cider_2015}, and BERTScore~\cite{zhang_bertscore_2020}. These metrics were used to assess different aspects of report quality, including lexical overlap, phrase-level agreement, and contextual similarity between generated and reference reports. The proposed model was compared with two representative baselines. PET2REP~\cite{zhang_pet2rep_2025} was selected because it is a PET/CT-based report generation model and therefore provides the most directly relevant comparison. MedVL-SAM2~\cite{xing_medvl-sam2_2026} was also included as a CT-only multimodal baseline, given the limited number of publicly available PET-specific report generation models. This comparison was used to assess whether PET/CT-specific visual-language training improves report generation beyond a general 3D CT vision-language framework. All models were evaluated under the same inference setting.
Table~\ref{tab:report_generation_comparison} summarizes the quantitative results, and qualitative examples are shown in Fig.~\ref{report_generation_fig}. As shown in Table~\ref{tab:report_generation_comparison}, the proposed model achieved the best performance across all evaluated metrics. Compared with PET2REP, the proposed model improved ROUGE from 16.73 to 23.42, BLEU from 27.40 to 36.17, METEOR from 16.98 to 19.62, CIDEr from 0.53 to 3.13, and BERTScore from 84.19 to 86.13. Larger improvements were observed relative to MedVL-SAM2, particularly for BLEU, METEOR, and CIDEr. These results suggest that PET/CT-specific multimodal training improves the generation of reports that are more closely aligned with reference PSMA PET/CT interpretations. 

\begin{table}[t]
\centering
\footnotesize
\caption{Comparison of MedVL-SAM2, PET2REP, and the proposed model using ROUGE, BLEU, METEOR, CIDEr, and BERTScore. Values are reported as mean [95\% CI].}
\label{tab:report_generation_comparison}

\renewcommand{\arraystretch}{1.15}
\setlength{\tabcolsep}{3pt}

\begin{tabular}{lccc}
\toprule
\textbf{Metric} &
\textbf{MedVL-SAM2} &
\textbf{PET2REP} &
\textbf{Proposed Model} \\
\midrule

\textbf{ROUGE}
& $12.4605\,[12.3097,\,12.6281]$
& $16.7270\,[16.4854,\,16.9981]$
& $\mathbf{23.4173\,[22.7848,\,24.0766]}$ \\

\textbf{BLEU}
& $9.1161\,[8.6978,\,9.5715]$
& $27.3971\,[26.9668,\,27.8269]$
& $\mathbf{36.1738\,[35.3247,\,37.0192]}$ \\

\textbf{METEOR}
& $7.6891\,[7.5851,\,7.7998]$
& $16.9818\,[16.7829,\,17.1689]$
& $\mathbf{19.6153\,[19.2467,\,19.9958]}$ \\

\textbf{CIDEr}
& $0.0151\,[0.0071,\,0.0571]$
& $0.5282\,[0.3971,\,0.6689]$
& $\mathbf{3.1294\,[1.8938,\,4.6403]}$ \\

\textbf{BERTScore}
& $82.6387\,[82.5830,\,82.7003]$
& $84.1922\,[84.1096,\,84.2757]$
& $\mathbf{86.1251\,[86.0005,\,86.2579]}$ \\

\bottomrule
\end{tabular}

\end{table}

\subsection{VQA}
\begin{figure}[htb]
\includegraphics[trim=0cm 0cm 0cm 0cm, clip, width=\textwidth]{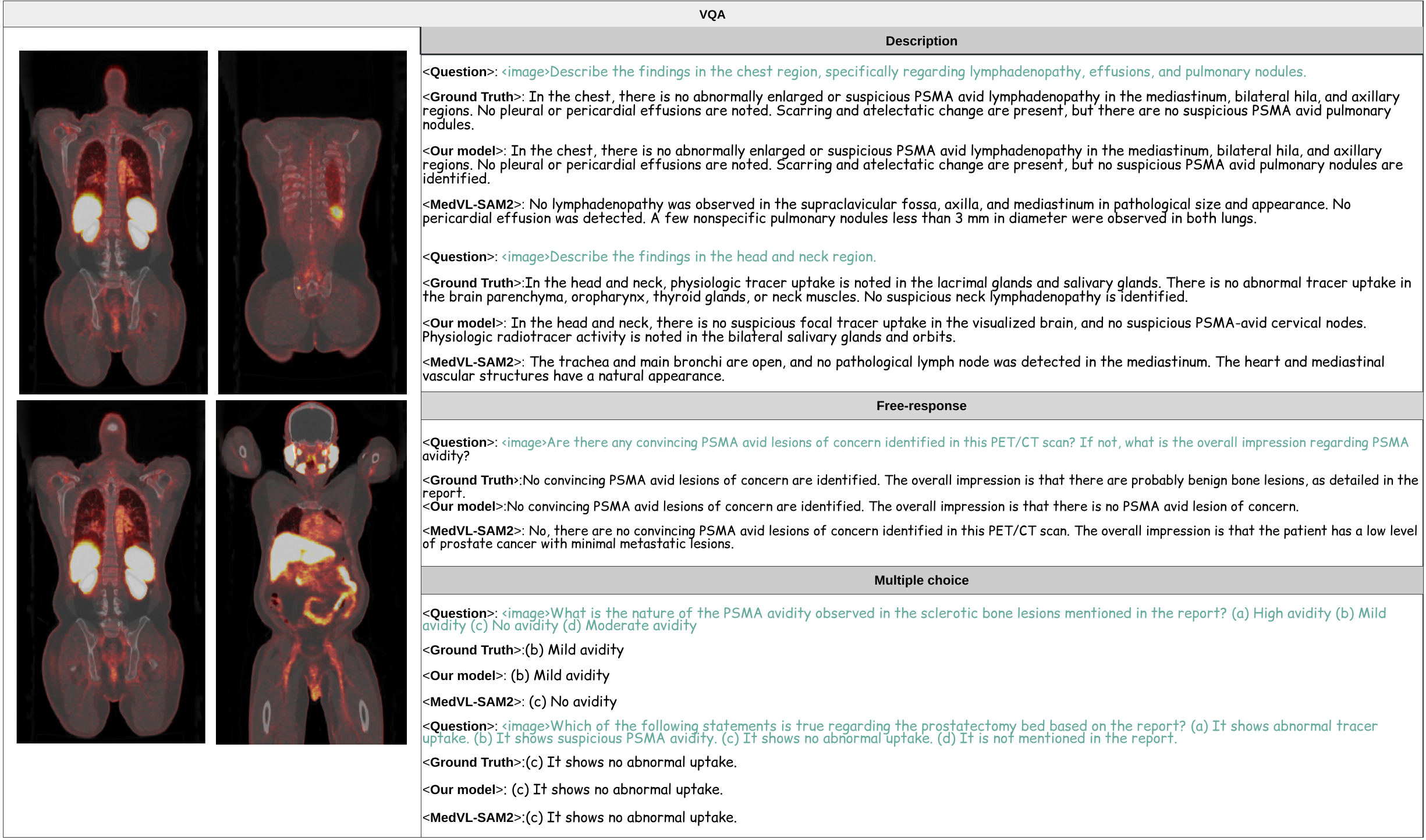}
\caption{Qualitative VQA results for 3 task types: description, free-response, and multiple-choice questions. Representative coronal slices from fused PET/CT images are displayed on the left, with the corresponding question, model response, and reference answer shown on the right. }
\label{vqa_fig}
\end{figure}

VQA was evaluated on the held-out PSMA PET/CT test set using three question formats: description, free-response, and multiple-choice questions. Description and free-response questions were treated as open-ended VQA tasks, in which the model generated text answers based on the PET/CT image and the input question. Multiple-choice questions were treated as closed-ended VQA, in which the model selected one answer from predefined options. These tasks were designed to assess whether the model could interpret clinically relevant PSMA PET/CT findings, including tracer uptake patterns, anatomic localization, disease extent, and impression-level diagnostic information. For open-ended VQA, performance was evaluated using ROUGE, BLEU, METEOR, CIDEr, and BERTScore, and for multiple-choice VQA, accuracy was used as the evaluation metric. Because few PET/CT-specific VQA models are publicly available, MedVL-SAM2 was used as the comparison model. Although MedVL-SAM2 was developed for CT-based multimodal analysis, it provides a relevant 3D medical vision-language baseline for evaluating whether PET/CT-specific training improves multimodal reasoning in this setting. As shown in Table~\ref{tab:vqa_comparison}, the proposed model outperformed MedVL-SAM2 across all VQA tasks and metrics. For description-based VQA, the proposed model improved ROUGE from 14.07 to 23.99, BLEU from 15.01 to 32.40, METEOR from 6.86 to 17.74, CIDEr from 1.88 to 18.82, and BERTScore from 84.86 to 87.45. For free-response VQA, larger gains were observed, with ROUGE increasing from 18.91 to 33.20 and CIDEr from 17.48 to 91.55. In the multiple-choice setting, accuracy increased from 70.89\% to 83.60\%. These results indicate that PET/CT-specific multimodal training improves both open-ended answer generation and closed-ended diagnostic question answering. The consistent improvement across description, free-response, and multiple-choice tasks suggests better alignment between 3D PSMA PET/CT findings and clinically oriented text responses, particularly for questions requiring interpretation of lesion distribution and disease status. Qualitative examples are shown in Fig.~\ref{vqa_fig}.

\begin{table}[t]
\centering
\footnotesize
\caption{Comparison of MedVL-SAM2 and the proposed model across VQA tasks, including description, free-response, and multiple-choice questions. Values are reported as mean [95\% CI], with confidence intervals estimated using bootstrap resampling.}
\label{tab:vqa_comparison}

\renewcommand{\arraystretch}{1.15}

\begin{tabular*}{\columnwidth}{@{\extracolsep{\fill}}llcc@{}}
\toprule
\textbf{Task} &
\textbf{Metric} &
\textbf{MedVL-SAM2} &
\textbf{Proposed Model} \\
\midrule

\multirow{5}{*}{Description}
& ROUGE
& $14.07\,[13.88,\,14.26]$
& $\mathbf{23.99\,[23.62,\,24.35]}$ \\

& BLEU
& $15.01\,[14.51,\,15.27]$
& $\mathbf{32.40\,[31.55,\,32.94]}$ \\

& METEOR
& $6.86\,[6.75,\,6.97]$
& $\mathbf{17.74\,[17.48,\,17.97]}$ \\

& CIDEr
& $1.88\,[1.60,\,2.22]$
& $\mathbf{18.82\,[16.76,\,21.11]}$ \\

& BERTScore
& $84.86\,[84.79,\,84.92]$
& $\mathbf{87.45\,[87.38,\,87.53]}$ \\

\midrule

\multirow{5}{*}{Free response}
& ROUGE
& $18.91\,[18.52,\,19.27]$
& $\mathbf{33.20\,[32.50,\,33.89]}$ \\

& BLEU
& $18.94\,[18.32,\,19.38]$
& $\mathbf{34.91\,[34.30,\,35.54]}$ \\

& METEOR
& $10.19\,[9.96,\,10.41]$
& $\mathbf{22.43\,[21.89,\,22.95]}$ \\

& CIDEr
& $17.48\,[15.70,\,19.27]$
& $\mathbf{91.55\,[85.33,\,98.04]}$ \\

& BERTScore
& $86.90\,[86.81,\,86.98]$
& $\mathbf{89.46\,[89.33,\,89.58]}$ \\

\midrule

Multiple choice
& Accuracy
& $70.89\,[69.29,\,72.39]$
& $\mathbf{83.60\,[82.39,\,84.87]}$ \\

\bottomrule
\end{tabular*}

\end{table}

\subsection{Lesion Segmentation}
\begin{figure}[htb]
\includegraphics[trim=0cm 4cm 0cm 0cm, clip, width=\textwidth]{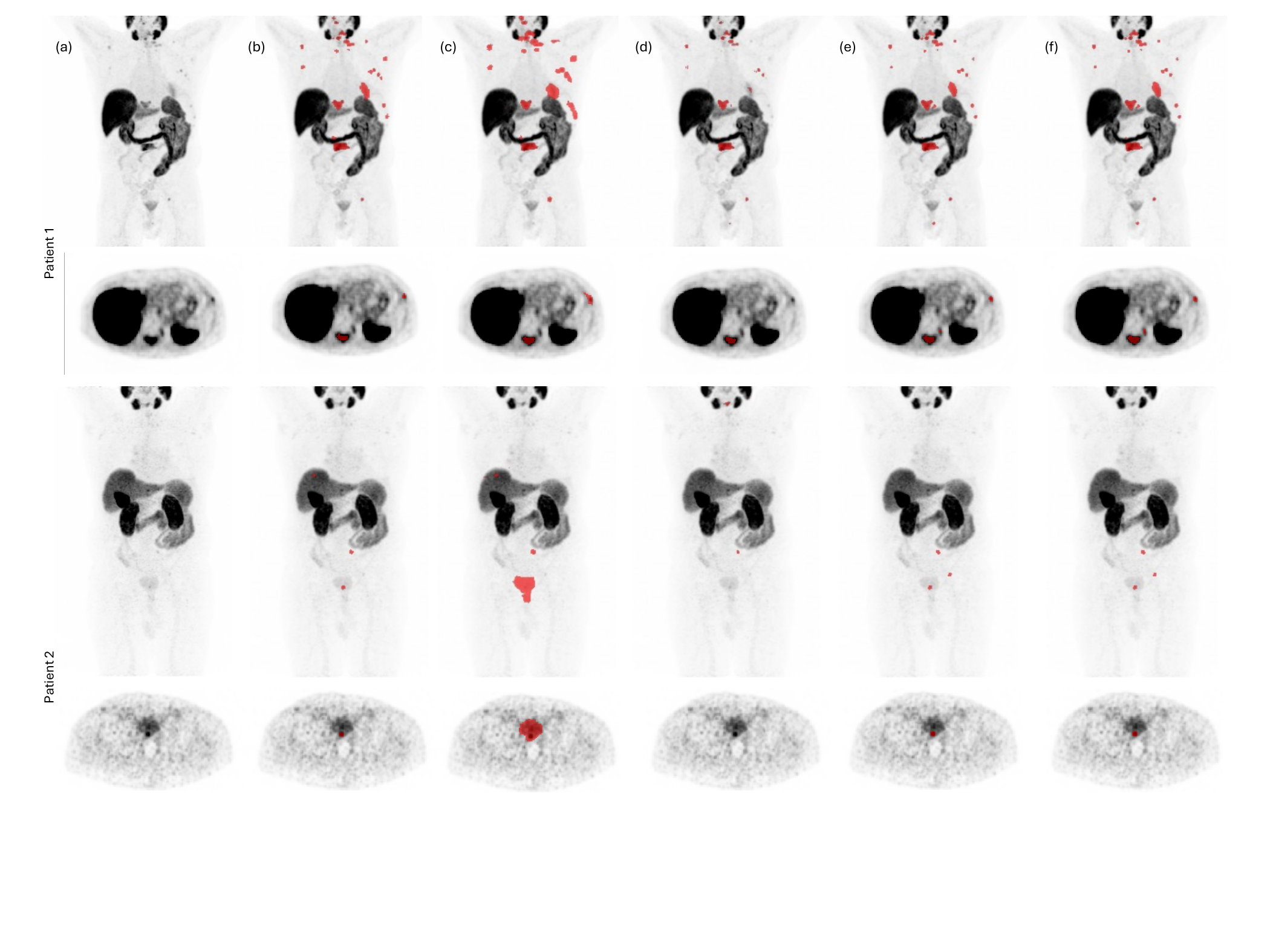}
\caption{Qualitative visualization of PSMA-avid lesion segmentation. Representative maximum-intensity-projection PET images and axial slices are shown for 2 different cases. (A) Original image, (B) reference lesion mask, (C) SegAnyPET, (D) nnUNet, (E) proposed model with semantic segmentation, and (F) proposed model with language-guided segmentation. Red overlays indicate segmented lesion regions.} 
\label{segmentation_fig}
\end{figure}

Lesion segmentation was evaluated on the PSMA PET/CT test set using case-level Dice coefficient and lesion-level F1 score. Dice was used to quantify voxel-level overlap with the reference lesion masks, and F1 assessed clinically relevant detection errors. F1 was computed under three different settings: (i) "Any overlap", where the prediction lesion was considered true positive if it overlaps with the ground truth target; (ii)"SUVmax match", where the perdiction lesion was considered true positive if it overlaps the SUVmax voxel of the reference lesion; (iii)"Dice$>$0.5", where the prediction lesion was considered true positve if the lesion-level Dice score was greater than 0.5. A predicted lesion that failed to fit the criteria was considered a false positive, and an unmatched ground truth lesion was considered a false negative. Bootstrap random sampling with replacement was used to calculate 95\% confidence intervals for each metric.  The proposed model was evaluated under two prompting settings: semantic segmentation, in which the instruction explicitly requested segmentation of PSMA-avid lesions, and language-guided segmentation, in which the model was guided by a natural language-based prompt. We compared the proposed model with SegAnyPET~\cite{zhang_seganypet_nodate} and nnUNet. SegAnyPET was included as a promptable PET segmentation baseline, whereas nnUNet was used as a strong supervised 3D medical segmentation baseline for lesion segmentation. This comparison allowed us to assess whether the proposed multimodal framework could preserve the segmentation strength of a task-specific model while enabling language-guided lesion localization. Quantitative results are summarized in Table~\ref{tab:segmentation_comparison}. The proposed model achieved the highest Dice scores in both semantic and language-guided settings, with Dice values of 0.5858 and 0.5875, respectively, outperforming SegAnyPET (0.4653) and nnUNet (0.5498). Similar improvements were observed for lesion-level detection. Under the "Any overlap" criterion, the proposed model achieved F1 scores of 0.7354 and 0.7356 for semantic and language-guided segmentation, respectively. Under the "SUVmax match" criterion, the corresponding F1 scores were 0.7269 and 0.7257. Using the stricter criterion "Dice$>$0.5", the proposed model achieved F1 scores of 0.5817 and 0.5925, outperforming both SegAnyPET and nnUNet. The semantic and language-guided settings showed comparable Dice performance, suggesting that the model can maintain lesion segmentation accuracy when guided by natural-language prompts.  Qualitative examples in Fig.~\ref{segmentation_fig} further illustrate the ability of the proposed model to localize PSMA-avid lesions compared with baseline methods. 

\begin{table}[t]
\centering
\caption{Comparison of SegAnyPET, nnUNet, and the proposed model for lesion segmentation. Values are reported as mean [95\% CI].}
\label{tab:segmentation_comparison}

\scriptsize
\renewcommand{\arraystretch}{1.15}
\setlength{\tabcolsep}{1.5pt}

\begin{tabular}{@{}lcccc@{}}
\toprule
\textbf{Model}
& \textbf{Dice}
& \shortstack{\textbf{F1}\\\textbf{SUVmax}}
& \shortstack{\textbf{F1}\\\textbf{Overlap}}
& \shortstack{\textbf{F1}\\\textbf{Dice $>$ 0.5}} \\
\midrule

SegAnyPET
& \shortstack{$0.4653$\\$[0.4327,0.4980]$}
& \shortstack{$0.6517$\\$[0.6126,0.6933]$}
& \shortstack{$0.6648$\\$[0.6278,0.7070]$}
& \shortstack{$0.2844$\\$[0.2572,0.3132]$} \\

nnUNet
& \shortstack{$0.5498$\\$[0.4962,0.6012]$}
& \shortstack{$0.6369$\\$[0.5911,0.6780]$}
& \shortstack{$0.6526$\\$[0.6068,0.6924]$}
& \shortstack{$0.4923$\\$[0.4393,0.5416]$} \\

\shortstack[l]{Proposed\\(Semantic)}
& \shortstack{$0.5858$\\$[0.5399,0.6293]$}
& \shortstack{$\mathbf{0.7269}$\\$\mathbf{[0.6876,0.7596]}$}
& \shortstack{$0.7354$\\$[0.6971,0.7694]$}
& \shortstack{$0.5817$\\$[0.5307,0.6282]$} \\

\shortstack[l]{Proposed\\(Language-guided)}
& \shortstack{$\mathbf{0.5875}$\\$\mathbf{[0.5426,0.6314]}$}
& \shortstack{$0.7257$\\$[0.6861,0.7604]$}
& \shortstack{$\mathbf{0.7356}$\\$\mathbf{[0.6961,0.7701]}$}
& \shortstack{$\mathbf{0.5925}$\\$\mathbf{[0.5451,0.6371]}$} \\

\bottomrule
\end{tabular}

\end{table}

\section{Discussion}\label{sec3}
This study presents a unified PSMA PET/CT VLM for report generation, VQA, and instruction-conditioned lesion segmentation. The results suggest that PET/CT-specific multimodal training improves performance on clinically oriented language tasks, including report generation and VQA, while preserving 3D lesion localization via an nnUNet-style segmentation module. The primary contribution of this framework is the development of a single instruction-driven system that supports both clinical language tasks and voxel-level lesion segmentation for PSMA PET/CT analysis. The proposed framework can be clinically useful in several ways. First, report generation can provide structured preliminary drafts that summarize major PSMA PET/CT findings. Second, VQA enables interactive querying of disease presence, distribution, and lesion-related findings. Third, lesion segmentation provides 3D voxel-level output that supports semantic and interactive tumor localization, as well as quantitative assessment. 

Compared with existing PET-oriented VLMs, which have primarily focused on image-text alignment, report generation, or instruction-following tasks~\cite{zhang_pet2rep_2025, jiao_vision-language_2025, maqbool_petar_2025}, the proposed framework extends language-based PSMA PET/CT interpretation to include explicit 3D lesion segmentation. Improvements in the language tasks suggest that PSMA-specific vision–language training can better capture the joint PET and CT information than models trained predominantly on other imaging domains. This benefit of domain-specific training is consistent with the nature of PSMA PET/CT interpretation, because identifying clinically meaningful findings and formulating diagnostic impressions require tracer-specific knowledge of physiologic biodistribution, disease-related uptake patterns, and their anatomical correlates on CT.  In contrast to dedicated segmentation methods such as nnUNet and SegAnyPET, which directly optimize voxel-level predictions, the proposed framework provides an instruction-conditioned interface: the language model specifies the target through the [SEG] representation, while an nnUNet-style decoder performs dense volumetric prediction. Auxiliary TotalSegmentator-derived organ masks further provide anatomical context that may help distinguish lesions from physiologic uptake in organs with characteristic PSMA expression. The segmentation results therefore support the feasibility of coupling language-conditioned interaction with a dedicated 3D segmentation architecture.  Taken together, the report-generation, VQA, and segmentation experiments evaluate complementary aspects of PSMA PET/CT analysis. Report generation assesses the ability to summarize whole-body findings in clinically meaningful text, VQA evaluates the ability to respond to targeted questions about the examination, and segmentation evaluates voxel-level localization of PSMA-avid lesions. Integrating these capabilities within one architecture provides a common instruction-driven interface while preserving the distinct output required by each task.

Several limitations exist in this preliminary study. First, the report-generation and lesion-segmentation datasets were not paired at the case level, which limited direct supervision and evaluation of sentence-level lesion grounding. \textcolor{black}{Specifically, a lesion identified by the segmentation branch was not explicitly constrained to be mentioned in the generated report, and a lesion described in the generated report was not explicitly required to correspond to a predicted mask. Paired image--report--mask datasets will be essential for training and evaluating such lesion-level correspondence. In our future work, we will construct paired image-report-mask datasets to enable explicit training and evaluation of lesion-level correspondence.} Second, the VQA dataset was derived from clinical reports and may therefore inherit report-level omissions or interpretation bias. Third, the AutoPET PSMA dataset differed from the in-house cohort in institution, acquisition protocol, and tracer distribution, which may affect segmentation generalizability. Future work will focus on constructing paired PSMA PET/CT datasets that include clinical reports, lesion masks, lesion-level attributes, and prompt-specific annotations from the same examinations. Such datasets would enable direct supervision and more rigorous evaluation of the generated report. In addition, future work will include detailed error analysis and prospective reader studies to determine whether the proposed unified framework can improve reporting consistency, diagnostic support, and clinical workflow efficiency.

In summary, we developed a unified PSMA PET/CT VLM for report generation, VQA, and lesion segmentation in both semantic and language-conditioned settings. PET/CT-specific vision-language training improved report-generation and VQA performance compared with baseline models, while the 3D segmentation module enabled voxel-level localization of PSMA-avid lesions through both semantic and language-guided prompts. These findings demonstrate the feasibility of a unified framework for clinical language tasks and 3D lesion segmentation, which may support more structured, interactive, and interpretable PSMA PET/CT analysis.

\section{Methods}\label{sec4}
\subsection{Dataset}
\subsubsection{PSMA PET/CT datasets with radiology reports}
We retrospectively collected 5,747 whole-body PSMA PET/CT examinations acquired with tracer 18F-DCFPyL between 2022 and 2025. This retrospective study was approved by the University of Florida Institutional Review Board with a waiver of informed consent. For spatial standardization, PET and CT images were resampled to a uniform spacing of 2.73 × 2.73 × 2.8 mm. After resampling, all examinations were cropped to a standardized anatomical field of view ending at the upper thighs and then padded or cropped to a fixed matrix size of 256 × 256 × 400. PET intensities were converted to standardized uptake values (SUV) units. The corresponding clinical reports were converted into structured JSON files. Each JSON record included the study date, findings section, and impression section. This structure preserved the clinically relevant organization of the original reports while enabling paired image-text training and downstream evaluation.

\subsubsection{VQA data construction}

\noindent We constructed the VQA dataset from the paired PSMA PET/CT examinations and structured radiology reports, following the report-derived VQA generation strategy used in CT-CHAT~\cite{hamamci_generalist_2026}. For each case, the de-identified findings and impression sections were used as the source text for question-answer generation. To reflect clinically relevant PSMA PET/CT interpretation, the generated questions focused on disease presence, disease distribution, local recurrence or prostate bed involvement, pelvic and extrapelvic nodal disease, and metastases. For each case, we generated three free-response questions, three multiple-choice questions, and three description-style questions using the Gemini-2.5-Pro API~\cite{comanici_gemini_2025}. Free-response questions were designed to extract concise diagnostic answers from the reports. Multiple-choice questions included one correct answer and plausible distractors. Description questions prompted short summaries of the major PET/CT findings or disease distribution. To improve data quality, VQA generation was performed using a two-step iterative procedure. In the first step, the language model generated candidate question-answer pairs from the structured report. In the second step, the model was prompted to verify whether each answer was directly supported by evidence in the report and to identify the corresponding supporting text. Question-answer pairs were retained only if they were correctly formatted, internally consistent, and supported by explicit report evidence. Unsupported, ambiguous, or malformed entries were discarded or regenerated. The pipeline of the data curation process is shown in Fig.~\ref{model_architecture}. C. The dataset was split at the case level to prevent information leakage between training, validation, and testing sets.  After quality control, the dataset comprised 4,023 training cases, 574 validation cases, and 1,150 testing cases. The training set contained 12,033 description, 11,772 multiple-choice, and 12,033 free-response questions; the validation set contained 1,722 questions for each VQA type; and the testing set contained 3,442 description, 3,390 multiple-choice, and 3,442 free-response questions.

\subsubsection{AutoPET PSMA lesion segmentation dataset}

The segmentation component was developed using the PSMA subset from the autoPET challenge~\cite{dexl_autopet3_2026}. This dataset contains 597 whole-body PSMA PET/CT studies from 378 male patients with suspected or diagnosed prostate cancer, acquired at LMU University Hospital Munich between 2014 and 2022. The lesion label was defined as a binary voxel-level mask of PSMA-avid tumor lesions, manually annotated in 3D on PET images. All PET/CT images and lesion masks were processed using the same spatial normalization pipeline as the in-house PSMA dataset. This dataset was used for semantic tumor segmentation training and quantitative evaluation of lesion-level segmentation performance. The dataset was split at the patient level into training, validation, and testing sets of 419, 59, and 119 studies, respectively.

\subsubsection{TotalSegmentator-derived organ masks}

To provide auxiliary anatomical supervision, physiologic uptake-related organ masks were generated from the CT images using TotalSegmentator~\cite{wasserthal_totalsegmentator_2023}. Ten organ or anatomic structure groups were selected based on their common physiologic or tracer-dependent uptake patterns, which may confound lesion detection on PSMA PET/CT. These included the kidneys, urinary bladder, liver, spleen, brain, lungs, heart, prostate, stomach, and salivary glands. For the salivary glands, parotid and submandibular gland masks were grouped into a single salivary gland category. For bilateral structures, left and right masks were merged into one organ class unless otherwise specified. These masks were used as auxiliary pseudo-labels for organ-aware supervision, helping the segmentation branch detect normal biodistribution patterns and distinguish suspicious lesions from physiologic uptake.

\subsection{Network architecture}
\subsubsection{Overview}
\textcolor{black}{3D medical vision-language tasks cover a broad spectrum, including report generation, VQA, and segmentation. To support all these tasks within a single model, we introduce a unified formulation that accommodates both text-centric and mask-centric predictions. Given a 3D PET/CT volume $\mathcal{I}_{in}\in\mathbb{R}^{2\times{X}\times{Y}\times{Z}}$ with dimension of $X\times{Y}\times{Z}$ and a sequence of text tokens $\mathcal{T}_{in}\in \mathbb{R}^{N\times{D}}$ with size of $N\times{D}$, the unified multi-task problem can be formulated as
\begin{equation}
\left(\mathcal{T}_{out},\mathcal{M}_{out}\right) = \mathcal{F}\left(\Theta;\mathcal{I}_{in},\mathcal{T}_{in}\right),
\end{equation}
where $\mathcal{F}(\cdot)$ denotes the unified 3D medical VLM, $\Theta=\{\theta_{vlb}, \theta_{seg}\}$ denotes the trainable parameters including the vision-language backbone parameters $\theta_{vlb}$ and the segmentation module parameters $\theta_{seg}$, $\mathcal{T}_{out}$ represents the output text tokens, and $\mathcal{M}_{out}$ is the output segmentation mask. The output text tokens  $\mathcal{T}_{out}$ are decoded by the tokenizer to generate the text output. }

The proposed model adopts a LLaVA-style multimodal framework to support PSMA PET/CT report generation, VQA, and lesion segmentation within a unified instruction-following architecture. The network consists of four main components: a 3D PET/CT image encoder, a multimodal projection layer, a large language model (LLM), and a language-conditioned nnUNet-style 3D segmentation module. The overall architecture is shown in Fig.~\ref{model_architecture} (B). The image-language branch encodes 3D PET/CT scans into visual tokens and aligns them with the language model embedding space. The LLM then generates free-text outputs, including radiology reports and answers to diagnostic questions. For segmentation tasks, a special [SEG] token is used to condition voxel-level mask prediction on the segmentation instruction. The final hidden-state representation of this token is projected into the segmentation feature space and fused with PET/CT features through a cross-attention module. This design allows the segmentation output to be conditioned on the text instruction while preserving the strong 3D localization capability of an nnUNet-style encoder-decoder network.

\subsubsection{Vision-language alignment}
Given an input PET/CT image, a 3D CLIP-based vision encoder $E_{vc}$ is used to extract image features \textcolor{black}{$v_{in}$, denoted as $v_{in}=E_{vc}(\mathcal{I}_{in})\in\mathbb{R}^{n \times d}$, where $n$ is the number of visual tokens and $d$ is the token dimension.} PET and CT images are provided as two channels to provide both molecular uptake information and anatomical context. Directly passing all 3D visual tokens into the language model is computationally inefficient because 3D patch embedding produces substantially longer token sequences than conventional 2D image inputs. To reduce this burden while preserving spatial information, an adaptive MLP-Mixer projection module~\cite{xin_med3dvlm_2025} is used to compress the visual tokens and align them with the language model embedding space. Given the visual tokens, the MLP-Mixer projection layer performs both channel and token mixing, producing language-compatible visual tokens that allow the model to reduce the visual token length while maintaining global context. \textcolor{black}{ Each MLP-Mixer layer contains two fully connected sublayers with nonlinearities applied independently to each row of the input tensor: the first performs channel mixing, and the second performs token mixing. Omitting layer indices, the mixer layer is formulated as
\begin{equation}
    \begin{split}
    U^T =  W_2\sigma(W_1\ \text{LayerNorm}(v_{in}^T)),\\
    T_v = W_4\sigma(W_3\ \text{LayerNorm}(U)),\\
    \end{split}
    \label{eq:mixer}
\end{equation}
where $\sigma$ denotes the GELU activation function, $U\in\mathbb{R}^{\hat{n}\times d}$ and $T_v \in \mathbb{R}^{\hat{n} \times \hat{d}}$ are the visual tokens aligned with the LLM's text embedding dimensions ( $\hat{n}=512$ and $\hat{d}=2048$ in our case). The computational complexity of the MLP-Mixer layer is linear with respect to the number of input patches, which enables efficient compression of 3D volumetric embeddings while preserving spatial information. }This design is more efficient than directly using a standard linear projection on the full 3D token sequence. The projected visual tokens are concatenated with the text instruction tokens and then passed to the language model, which generates radiology reports, answers clinical questions, and produces task-specific output tokens for segmentation. LoRA layers (rank=128) are applied to the language backbone to enable parameter-efficient fine-tuning while preserving the general language capability of the pretrained model. \textcolor{black}{The overall process is formulated as: $T_{out} = LLM(T_{in}+T_v)$, where $T_{{in}}$ denotes the tokenized text prompt and $T_v$ denotes the visual tokens produced by the projection layer.}

\subsubsection{Semantic and language-guided segmentation module}
The segmentation branch follows an nnUNet-style 3D encoder-decoder design~\cite{isensee_nnu-net_2021}. Unlike the 3D vision encoder, which is optimized for global image-text alignment, the segmentation branch is optimized for dense 3D voxel-level prediction. It receives the preprocessed PET/CT image and extracts multi-resolution features through a residual encoder \textcolor{black}{$f_{enc}$: $\mathcal{Z}=\{z_1,...,z_{s}\}, z_s\in\mathbb{R}^{C_s \times H_s \times W_s \times D_s}$, where $S$ is the number of scales}. These features are then decoded with skip connections to generate the final 3D lesion mask\textcolor{black}{: $y_s=f_{dec,s}\left (y_{s-1}, z_s\right)$.} For segmentation tasks, the instruction prompt asks the model to identify the target lesion or structure and produce a special [SEG] token. The final-layer hidden representation of this token is extracted from the language model and projected into the segmentation feature space, \textcolor{black}{denoted as \texttt{[SEG]}$_{hs}$}. This representation encodes the target specified by the text instruction, such as PSMA-avid lesion segmentation or a language-guided segmentation request. To condition mask prediction on the text instruction, the projected [SEG] representation is fused with bottleneck segmentation features through a cross-attention module, denoted as \textcolor{black}{$\tilde{z}_s = CA(\texttt{[SEG]}_{hs}, z_s)$, where \texttt{[SEG]}$_{hs}$ is the query and $z_s$ is the key and value for the cross attention module}. This fusion allows the segmentation branch to emphasize image features relevant to the text-specified target before mask decoding. The resulting language-conditioned multi-resolution features are passed to the decoder to produce the final 3D probability map, which is subsequently converted into a binary lesion mask. This design combines language-conditioned target specification with a specialized medical segmentation backbone. 
\subsection{Implementation details}
The model was trained using a four-stage strategy to progressively establish PET/CT visual encoding, vision-language alignment, task-specific language generation, and voxel-level segmentation. First, the 3D PET/CT vision encoder was trained using contrastive learning~\cite{radford_learning_nodate} to align 3D image features with corresponding report-level text representations. This stage provided an imaging-specific initialization for encoding whole-body PSMA PET/CT images. The training process of the vision encoder is shown in Fig.~\ref{model_architecture}. A. Second, the vision-language projection layer was trained to map PET/CT visual tokens into the language model embedding space. During this stage, the vision encoder and language model were frozen, and only the projection module was optimized using report-generation and VQA samples with the autoregressive language modeling loss. Third, the VLM branch was fine-tuned for report generation and VQA. The PET/CT vision encoder, projection layer, and LoRA parameters of the language model were optimized, while the segmentation branch was not updated. This stage strengthened PET/CT-specific image-text reasoning before introducing voxel-level supervision. Finally, multitask tuning was performed using mixed instructions for report generation, VQA, semantic segmentation, and language-guided segmentation. The VLM branch, segmentation-token interface, cross-attention fusion module, and nnUNet-style segmentation branch were jointly optimized. Text-generation samples were supervised using autoregressive language modeling loss, while segmentation samples were supervised using a combination of cross-entropy and Dice losses. During testing, task-specific prompts were used for report generation, VQA, semantic segmentation, and language-guided segmentation. 
\section{Statistical analysis}
\textcolor{black}{Statistical analysis was performed to quantify the uncertainty and robustness of model performance. Bootstrap resampling was used to estimate 95\% confidence intervals for all evaluation metrics. For image-text retrieval, Recall@K was evaluated for both image-to-text and text-to-image retrieval. Report generation was evaluated using ROUGE, BLEU, METEOR, CIDEr, and BERTScore. VQA performance was evaluated using language-generation metrics for description and free-response questions and accuracy for multiple-choice questions. Lesion segmentation was evaluated using the case-level Dice coefficient and lesion-level F1 scores under the SUVmax-match, any-overlap, and Dice ($>$0.5) criteria.}

\section{Data availability}\label{sec5}
The institutional PSMA PET/CT imaging data and corresponding clinical reports used in this study are not publicly available because of patient privacy and institutional ethical restrictions. De-identified data may be made available to qualified researchers upon reasonable request to the corresponding author, subject to institutional approval and execution of an appropriate data-use agreement. The publicly available AutoPET dataset used for lesion-segmentation experiments can be accessed through its original repository: \url{https://autopet-iv.grand-challenge.org}. 

\section{Code availability}\label{sec6}
% The source code used for model training, inference, preprocessing, and evaluation will be made publicly available upon publication at GitHub at: \url{https://github.com/astlian9/PET_VLM}. Currently, the code and model weights can be downloaded from \href{https://www.dropbox.com/scl/fo/2fi0ejexo2sixmlev15hq/APGlNaQ7WKe4p3bvx3wSRQI?rlkey=qkkn1fx4c2cbor9o6afb8pbpb&st=fpmv7cjq&dl=0}{\underline{this link}}. 

% The code will include model configurations and instructions required to reproduce the principal experiments reported in this study. 
Model training was performed using the AdamW optimizer as implemented in PyTorch. For baseline comparisons, we used the CTCLIP model (hf-hub:\url{https://huggingface.co/datasets/ibrahimhamamci/CT-RATE}) via the HuggingFace hub. All experiments were implemented using PyTorch (v2.7) and Transformers (v4.42), and executed on a cluster equipped with eight NVIDIA B200 GPUs (180 GB each).
 
\section*{Acknowledgements}
This work was supported by NIH grants R01EB034692 and R01AG078250.

\section*{Author contributions}
Y.X. processed the data, developed the code, trained the models, performed the experiments, analyzed the results, prepared the figures, and drafted the manuscript. J.W., Y.Y., W.S., and C.Y. contributed to model and experimental design and interpretation of the results. T.P., Y.Y., and Z.Z. collected the data. Y.Z. contributed to the VQA construction pipeline. B.Y. and Z.Z. performed data processing. K.W., S.O., and Y.L. contributed to clinical task design and analysis of the results. K.G. guided the project,provided research support, and advised on technical details and overall direction.  All authors reviewed, revised, and approved the manuscript.

\section*{Competing interest}
The authors declare no competing interests.

\nolinenumbers
\bibliographystyle{sn-nature}
\bibliography{sn-bibliography}
% \bibliography{sn-bibliography}% common bib file

\end{document}